\documentclass{article}

\usepackage{arxiv}

\usepackage[utf8]{inputenc} 
\usepackage[T1]{fontenc}    
\usepackage{hyperref}       
\usepackage{url}            
\usepackage{booktabs}       
\usepackage{amsfonts}       
\usepackage{nicefrac}       
\usepackage{microtype}      
\usepackage{lipsum}		
\usepackage{graphicx}
\usepackage{natbib}
\usepackage{doi}

\usepackage{ mathrsfs }
\usepackage{placeins}
\usepackage{amsmath}
\usepackage{caption}
\usepackage{subcaption}
\usepackage{xcolor}

\usepackage[super]{nth}

\title{`Next Generation' Reservoir Computing: an Empirical Data-Driven Expression of Dynamical Equations in Time-Stepping Form}

\date{January 13, 2022}	

\author{{Tse-Chun Chen}\thanks{Corresponding author: Tse-Chun Chen, tse-chun.chen@noaa.gov}, \hspace{1mm} \href{https://orcid.org/0000-0002-5223-8307}{\includegraphics[scale=0.06]{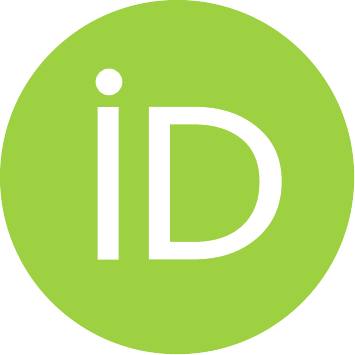}}\hspace{1mm} {Stephen G. Penny},  \hspace{1mm} {Timothy A. Smith} \\
	Cooperative Institute for Research in Environmental Sciences\\
	University of Colorado Boulder\\
	Boulder, CO 80309 \\
	Physical Sciences Laboratory\\
	National Oceanic and Atmospheric Administration\\
	Boulder, CO 80305
	\And
	{Jason A. Platt} \\
	Department of Physics\\
	University of California San Diego\\
	La Jolla, CA 92093 \\
}

\renewcommand{\shorttitle}{NVAR-RC: Empirical Data-Driven Time-Stepping Schemes }

\hypersetup{
pdftitle={NVAR-RC-time-stepping},
pdfsubject={q-bio.NC, q-bio.QM},
pdfauthor={Tse-Chun Chen},
pdfkeywords={Nonlinear Vector Autoregression, Reservoir Computing, Numerical Methods, Dynamic Identification},
}

\begin{document}
\maketitle

\begin{abstract}
Next generation reservoir computing based on nonlinear vector autoregression (NVAR) is applied to emulate simple dynamical system models and compared to numerical integration schemes such as Euler and the \nth{2} order Runge-Kutta. It is shown that the NVAR emulator can be interpreted as a data-driven method used to recover the numerical integration scheme that produced the data. It is also shown that the approach can be extended to produce high-order numerical schemes directly from data. The impacts of the presence of noise and temporal sparsity in the training set is further examined to gauge the potential use of this method for more realistic applications.
\end{abstract}


\section{Next generation reservoir computing base on nonlinear vector autoregression}
\subsection{Background}
Reservoir Computing (RC) has generated wide interest in learning time series of dynamical systems. A series of studies \citep{lu_reservoir_2017,pathak_model-free_2018,griffith_forecasting_2019,arcomano_machine_2020,platt_robust_2021} demonstrated the potential of RC in predicting short-term evolution and reproducing long-term climate attractors for dynamical systems with a wide range of complexity. Next generation RC has been proposed as an explanation of the success of standard RC \citep{gauthier_next_2021,bollt_explaining_2021,pyle_domain-driven_2021}. The new approach is also an attractive alternative to standard RC as it requires far less setup and training effort while often showing equal if not better performance. In contrast to the randomly initialized recurrent neural network in the standard RC, the essence of this new approach is to form a range of polynomial features from the input signal as a simplified reservoir. A readout operator is then identified to match these features to the desired output target as in the standard RC. Because this procedure is very similar to the construction of a classical vector autoregression model but with nonlinear products of the input variables as additional features, it is also referred to as nonlinear vector autoregression (NVAR) machine \citep{bollt_explaining_2021}. To reflect this nature of the method, we will use the name NVAR-RC throughout the paper. Note that we can include more than just the polynomials of the input variables to the features in NVAR-RC. This generalization will be discussed in later sections.

Here we loosely follow the notation of \cite{gauthier_next_2021}. Given an input data from a 2-variable system $\mathbf{X} = [x_0 \quad x_1]^T$, the NVAR-RC forms a vector of polynomial features $\mathbf{R}=[1  \quad  x_0 \quad x_1 \quad x_0^2 \quad x_0x_1 \quad...\quad x_0^p \quad x_0^{p-1}x_1 \quad ... \quad x_1^p  ]^T$ and uses a readout operator (or linear output layer) $\mathbf{W}_{out}$ to map the features to the desired output $\mathbf{Y}$. In the training phase, the $\mathbf{W}_{out}$ is determined such that the output closely approximates the desired $\mathbf{Y}$, and the procedure can be as simple as finding the solution of $\mathbf{Y}=\mathbf{W}_{out} \mathbf{R}$ through a least-square fit over a series of samples. A common practice is the use of a Tikhonov regularization so that the output layer is given by
\begin{equation} \label{eq:Wout1}
\mathbf{W}_{out}=\mathbf{YR}^T (\mathbf{RR}^T+\alpha\mathbf{I})^{-1},
\end{equation}
where $\mathbf{I}$ is the identity matrix. The Tikhonov parameter $\alpha > 0$ helps to avoid overfitting and decreases the condition number of ill-posed problems, with a tradeoff of introducing a tolerable amount of bias. Note that setting $\alpha > 0$ decreases the fit to data and hence degrades the result for the problems that are noise-free and not ill-posed. 

Note that the NVAR-RC setup is quite similar to the Sparse Identification of Nonlinear Dynamics (SINDy) algorithm developed by \cite{brunton_discovering_2016}. The major difference is that SINDy regress-fits the "functional library" ($\mathbf{R}$ in NVAR-RC) to the derivative of the measured variables ($\mathbf{Y}$ in NVAR-RC) as opposed to the variables in the next time step as in NVAR-RC. In this way, NVAR-RC can be viewed as the mapping version of SINDy. One advantage of NVAR-RC over SINDy is that the numerical approximation to derivatives from the data is not required.

The hyperparameters include the highest degree $p$ of the polynomial features and the number of time-lagged steps $t$ for the input. For simplicity, we set $\alpha=0$ since we mostly explore NVAR-RC in a perfect data scenario, and applying Tikhonov regularization appears to degrade the performance.

\subsection{Preconditioning}
Alternatively to the form given in equation \ref{eq:Wout1}, we can instead precondition the regression problem by solving the modified form
\begin{equation}  \label{eq:Wout2}
\mathbf{W}_{out}=\mathbf{Y}(\mathbf{SR})^T ((\mathbf{SR})(\mathbf{SR})^T)^{-1}\mathbf{S},
\end{equation}
where $\mathbf{S}$ is a diagonal matrix with positive scalar entries of the inverse of the maximum values of each feature in $\mathbf{R}$ over the training period. The preconditioning addresses numerical challenges with the inversion of $\mathbf{RR}^T$ in eqn. \ref{eq:Wout1}, and allows better estimation of each entry. We found this technique improves the performance of NVAR-RC in general, and is particularly helpful in avoiding large magnitude differences between features when including high-degree polynomial features, which tend to make $\mathbf{RR}^T$ ill-conditioned.

\section{Connecting the reservoir readout operator $\mathbf{W}_{out}$ to time-stepping dynamical equations}
\subsection{Deriving the time-stepping equations}
In this section, we introduce a framework that draws a direct connection between the NVAR-RC readout operator $\mathbf{W}_{out}$ and the underlying dynamical equations together with the specific numerical integration scheme used to generate the training data. In this study, we focus on two classic dynamical systems: the Lorenz 1963 model (L63) \citep{lorenz_deterministic_1963} and the Lorenz 1996 model (L96) \citep{lorenz_optimal_1998}. For simplicity, we show examples using the 1st order forward Euler scheme unless otherwise stated. 

Starting with the L63 system, its equations in differential form are:

\begin{equation}\label{eqn:l63}
\begin{aligned}
\frac{dx}{dt}=&\sigma (y-x)\\
\frac{dy}{dt}=&\rho x-y-xz\\
\frac{dz}{dt}=&xy-\beta z
\end{aligned},
\end{equation}

where $\sigma=10, \rho=28$, and $\beta=\frac{8}{3}$. These equations can be written in a time-stepping form by using an explicit numerical integration scheme and solving for the next time step. Given a general ordinary differential equation $\frac{du}{dt}=f(u,t)$, the Euler scheme yields $u_{n+1}=u_n + h f(u_n,t_n)$, where $h$ represents the discretized timestep at time index $n$. The Euler time-stepping form of Equation \ref{eqn:l63} can then be written as
\begin{equation} \label{eqn:l63_stepping}
\begin{bmatrix} x_{n+1} \\ y_{n+1} \\ z_{n+1} \end{bmatrix} =\begin{bmatrix} (1-h\sigma)x_n + h\sigma y_n \\ h\rho x_n + (1-h) y_n -h x_n z_n \\ 
(1-h\beta)z_n + h x_n y_n
\end{bmatrix}.
\end{equation}

\begin{figure}
    \centering
    \includegraphics[clip, trim=0cm 9cm 0cm 0cm, width=\textwidth]{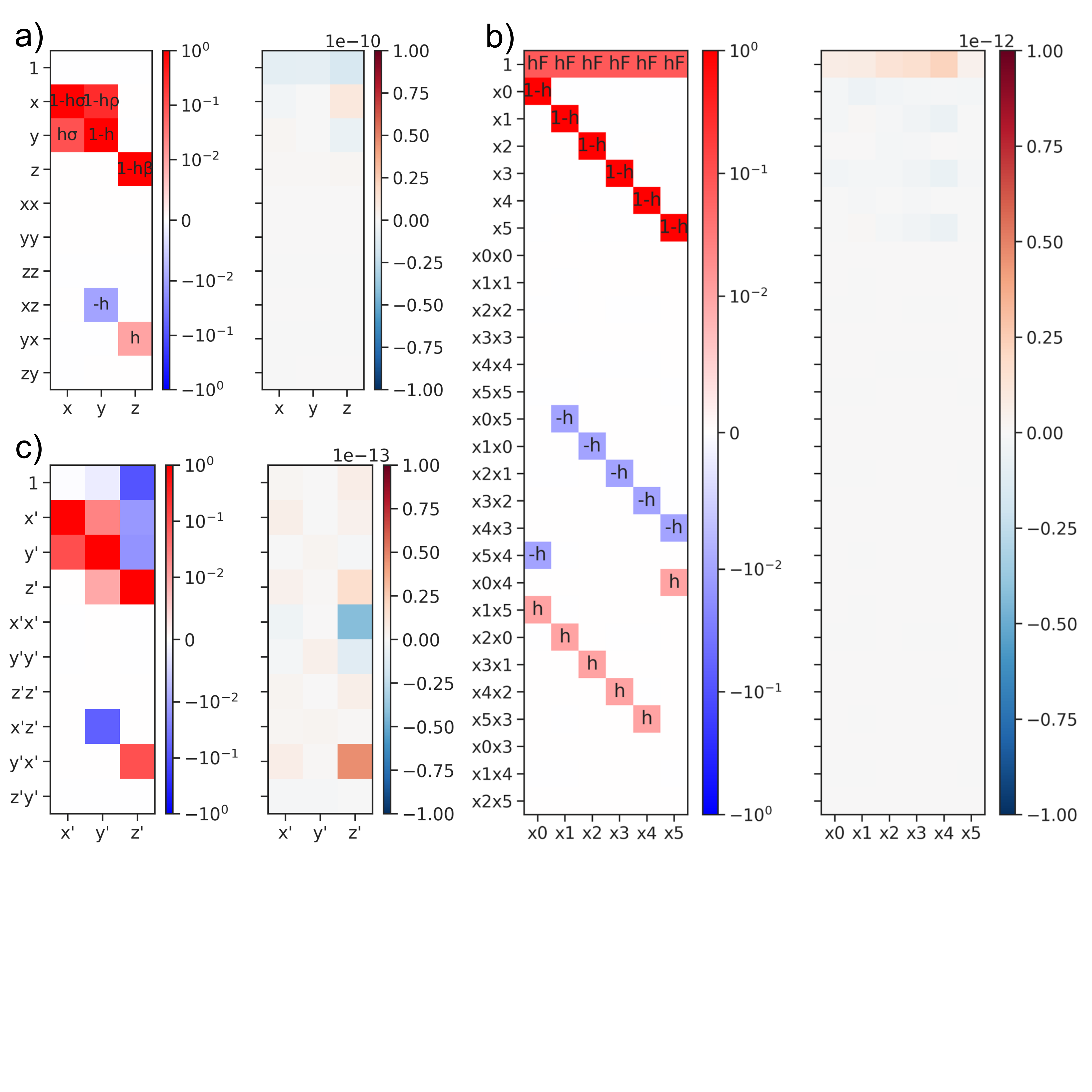}
    \caption{Visualization of the data-driven NVAR-RC readout matrices and the differences from their corresponding $\mathbf{W}_{Euler}$ matrices that were derived from the Euler time-stepping equations, for
(a) $\mathbf{W}_{out}^{L63}$ as computed for the L63 system, (b) $\mathbf{W}_{out}^{L96}$ as computed for the L96-6D system, and (c) $\mathbf{W}_{out}^{L63 norm}$ as computed for the normalized L63 system. The $\mathbf{W}_{Euler}$ entries are annotated on the corresponding grids for (a) L63 and (b) L96-6D systems. For clarity, the $\mathbf{W}_{Euler}$ for the normalized L63 is shown in Table \ref{tab:wout_l63_norm_Euler}. }
    \label{fig:wout_Euler_lag0}
\end{figure}

Assuming the right-hand side terms are a subset of the polynomial features in the NVAR-RC, then the corresponding $\mathbf{W}_{out}$ identifies the coefficients needed for the Euler numerical integration scheme. The corresponding coefficients for the L63 system using the Euler scheme can be derived from Equation {\ref{eqn:l63_stepping}} and are shown as the annotation in Figure \ref{fig:wout_Euler_lag0} (a).

The L96 model is governed by
\begin{equation} \label{eqn:l96}
\frac{dx_i}{dt}=(x_{i+1}-x_{i-2})x_{i-1}-x_i+F,
\end{equation}
where $i=1,...,N$ and $x_0=x_N, x_{-1}=x_{N-1}, x_{N+1}=x_0$, and $N\ge4$. In this study, we choose the system dimension $N=6$ and the external forcing constant $F=8$, which are common values for generating chaotic time series. We will refer to this system as L96-6D hereafter. Again, using the Euler scheme, Equation {\ref{eqn:l96}} can be rewritten in the time-stepping form
\begin{equation}
x_{i,n+1}=hF+(1-h)x_{i,n}+h(x_{i+1,n}-x_{i-2,n})x_{i-1,n},
\end{equation}
where $x_{i,n}$ is the state variable on $i$th grid at time step $n$. The corresponding $\mathbf{W}_{out}$ that contains the coefficients of the Euler scheme can be expressed as the annotation in Figure \ref{fig:wout_Euler_lag0} (b).

Note that the time-stepping form of the dynamical equations changes with the integration scheme used. The right-hand sides of both dynamical systems contain at most \nth{2} degree polynomial terms. When expanding to the time-stepping form using the \nth{1} order forward Euler scheme, each term on the right-hand side in equations \ref{eqn:l63} and \ref{eqn:l96} (i.e. flow rate form) can be found in the nonzero entries of the readout operator $\mathbf{W}_{out}$, which are also at most \nth{2} degree polynomials. For instance, one distinct difference between the two derived $\mathbf{W}_{out}$ readout operators is the contribution of the bias feature (\nth{0} degree). The L63 system has all zero entries for the bias features while the L96-6D has a uniform contribution of a constant $hF$ for all variables, which corresponds to the constant forcing term in the original dynamical equation. Higher-degree polynomial features and time-lags will appear in $\mathbf{W}_{out}$ when using a higher-order numerical integration scheme for the same dynamical systems. For clarity, the readout operator derived from the Euler time-stepping equations will be referred to as $\mathbf{W}_{Euler}$.

Having derived $\mathbf{W}_{Euler}$ for both systems, we now compare the NVAR-RC based $\mathbf{W}_{out}$ that is trained by performing a least-square fit to the data. Note that the Euler scheme is used to generate both the training and testing data, using a time integration step size of $0.01$ for both L63 and L96-6D. We use a training data length of 400 model time units (MTU). To obtain robust test results, we generated 100 samples distributed at different points over the attractor to create 100 different 25 MTU test datasets. The prediction skill metric is the valid prediction time (VPT) defined as the time $T$ when $$RMSE(T)=\sqrt{\sum_i^D[\frac{x_i^{true}(T)-x_i^{pred}(T)}{\sigma_i}]/D}>\epsilon,$$ where D is the dimension of the system, $\sigma_i$ is the standard deviation of each dimension over the training period, and $\epsilon=0.3$ is an arbitrary threshold for determining if the prediction remains close to the truth relative to the climatological variation.  

Figure {\ref{fig:wout_Euler_lag0}} shows the NVAR-RC $ \mathbf{W}_{out}$ for the L63 and L96-6D systems and their differences from their corresponding $ \mathbf{W}_{Euler}$ matrices as derived from equations (annotated). It is clear that the NVAR-RC reconstructs the dynamics in the time-stepping form with small errors. One could also observe that the error magnitudes are negatively correlated with the degree of the corresponding polynomial features. 

\begin{figure}
    \centering
    \includegraphics[clip, trim=0 7cm 0 0, width=\columnwidth]{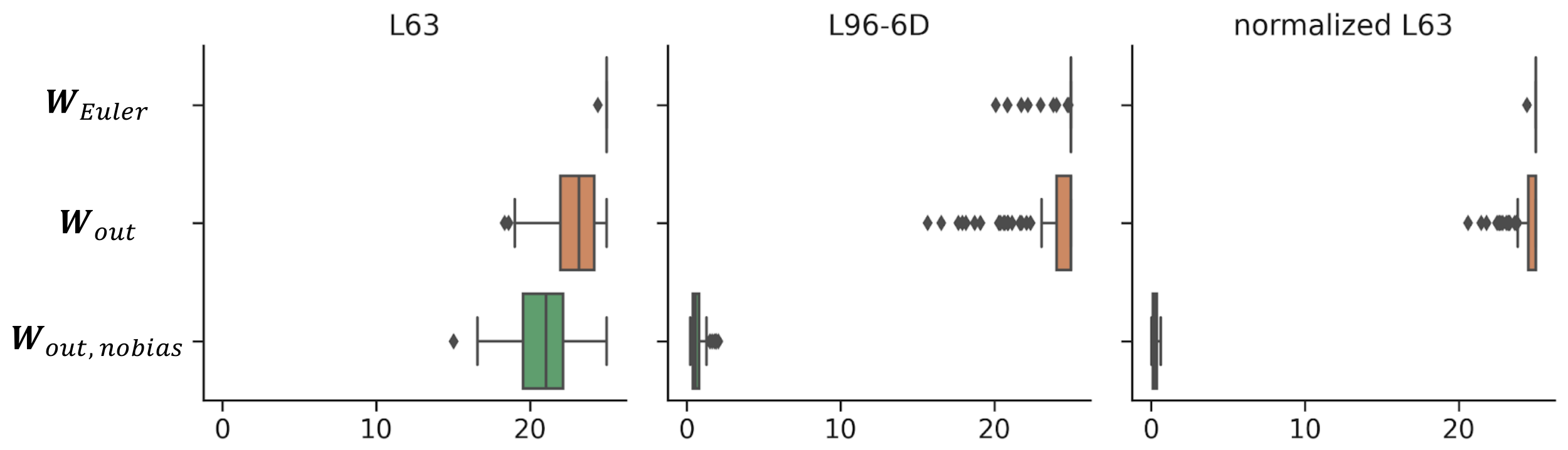}
    \caption{Comparisons of valid prediction time (VPT) from the derived $\mathbf{W}_{Euler}$ and NVAR-RC $\mathbf{W}_{out}$ with and without the \nth{0} degree bias feature on predicting L63(left), L96-6D (center), and normalized L63 (right). Note that using $\mathbf{W}_{Euler}$ makes almost perfect prediction up to the length of testing data (25 MTU or ~22 Lyapunov time for both L63 and L96-6D), while the NVAR-RC with the bias feature is also exceedingly skillful up to about 23 MTU. Excluding the bias feature in NVAR-RC degrades the performance significantly for the normalized L63 and L96-6D prediction, while performance is only slightly degraded for the L63 prediction.}
    \label{fig:vpt_Euler}
\end{figure}

Figure {\ref{fig:vpt_Euler}} compares the prediction performance using $\mathbf{W}_{Euler}$ as derived from equations against the NVAR-RC $ \mathbf{W}_{out}$ either with or without the bias feature. The prediction using  $\mathbf{W}_{Euler}$ is almost perfect for the entire 25 MTU test period for all the samples, while the NVAR-RC with the bias feature produces exceedingly skillful predictions up to about 23 MTU. This is because the $\mathbf{W}_{Euler}$ produces essentially the same set of equations as the Euler scheme that was used to generate the data (with some round-off errors). In a sense, the $\mathbf{W}_{Euler}$  is the correct answer for NVAR-RC but the small errors ($<10^{-10}$) in $\mathbf{W}_{out}$ accumulate over each iteration leading to the forecast divergence at 23 MTU. The linear regression of NVAR-RC effectively `discovers' the Euler time-stepping equations, with some estimation error. Excluding the bias feature degrades the performance significantly for L96-6D but remains a similar VPT for L63. This corresponds to the dynamical equations in the time-stepping form where the L96-6D presents a nonzero contribution from the bias feature that comes from the forcing $ F$ in the dynamical equations. This helps to explain the results of \citet{platt_practical_nodate} [\textit{in prep.}] who showed that including a bias term in the standard RC can produce dramatic improvements, especially for the L96 system.

\subsection{Impact of normalizing data}
Preconditioning the computation of $\mathbf{W}_{out}$ was found to be helpful in general and was particularly effective for incorporating higher-order polynomial features by decreasing the condition number. Normalizing data has a similar effect on NVAR-RC, but this also changes the time-stepping dynamical equation. A widely used normalization centers each variable to its time mean value and scales the standard deviation to one. A more general form of this transformation can be expressed by letting $ u=\overline{u}+k_uu'$ , where $ u,\ \overline{u},\ k_u$ , and $ u'$  are the original variable, centering constant, scaling factor, and the normalized variable. Using the Lorenz 63 system as an example, we can derive again the time-stepping equations by applying such a transformation to equation {\ref{eqn:l63_stepping}}. The resulting $ \mathbf{W}_{Euler}^{L63norm}$ for the normalized Lorenz 63 system formulated with the Euler scheme is shown in Table \ref{tab:wout_l63_norm_Euler}. Note that there are more non-zero entries in this new $ \mathbf{W}_{Euler}$ compared to the non-normalized form. It is generally the case that normalization produces more nonzero entries if there are \nth{2} (or higher) degree polynomials on the right-hand side of the dynamical equation, and the centering constant $\overline{u}$ is nonzero. Among the differences, the most obvious new entries are the ones associated with the bias features, showing that additional bias features are now required after normalization. From the perspective of dynamics identification, it may be preferable to apply the preconditioning technique over standard data normalization to avoid introducing these new complicating terms.

\begin{table}
\centering
\begin{tabular}{c c c c} 
 \hline
       & $x_{n+1}'$ & $y_{n+1}'$ & $z_{n+1}'$ \\ [0.5ex] 
 \hline
 1           & $h\sigma k_x(\bar{y}-\bar{x})$ & $hk_y(\rho\bar{x}-\bar{y}-\bar{x}\bar{z})$ & $hk_z(\bar{x}\bar{y}-\beta \bar{z})$ \\ 
 $x_n'$      & $1-h\sigma$ & $h(\rho-\bar{z})k_y/k_x$ & $h\bar{y}k_z/k_x$ \\
 $y_n'$      & $h\sigma k_x/k_y$   & $1-h$   & $h\bar{x}k_z/k_y$ \\
 $z_n'$      & 0           & $-h\bar{x}k_y/k_z$       & $1-h\beta$ \\
 $x_n'^2$    & 0           & 0       & 0 \\
 $y_n'^2$    & 0           & 0       & 0 \\
 $z_n'^2$    & 0           & 0       & 0 \\
 $x_n' z_n'$  & 0           & $-hk_y/k_xk_z$    & 0 \\
 $x_n' y_n'$  & 0           & 0       & $hk_z/k_yk_x$ \\
 $y_n' z_n'$  & 0           & 0       & 0 \\
 \hline
\end{tabular}
\caption{\label{tab:wout_l63_norm_Euler} $\mathbf{W}_{out}^{L63 norm}$ of the normalized L63 using the Euler scheme}
\end{table}

It is worth noting here that the extra nonzero entries of the bias features that appear after normalizing L63 are consistent with our experience on training standard RC. We found that the normalized L63 and L96 systems are very difficult to learn by standard RC unless the bias feature is included in the RC formulation \citep[][\textit{in prep.}]{platt_practical_nodate}. One explanation could be that the linear features are passed along through the isolated nodes and the nonlinear features are generated by the recurrent connections, but there is no way for the randomized reservoir to provide a bias feature. This connection between the standard RC and the NVAR-RC may point to directions in designing reservoir connections for improving the standard RC or a hybrid approach that provides features including polynomials, special functions, and transformations to the recurrent reservoir \citep[e.g.,][]{lin_fourier_2021}. We will also show an example of including special functions for NVAR-RC in a later section.

\subsection{Cross-validation on datasets using different integration scheme}
\begin{figure}
    \centering
    \includegraphics[width=\columnwidth]{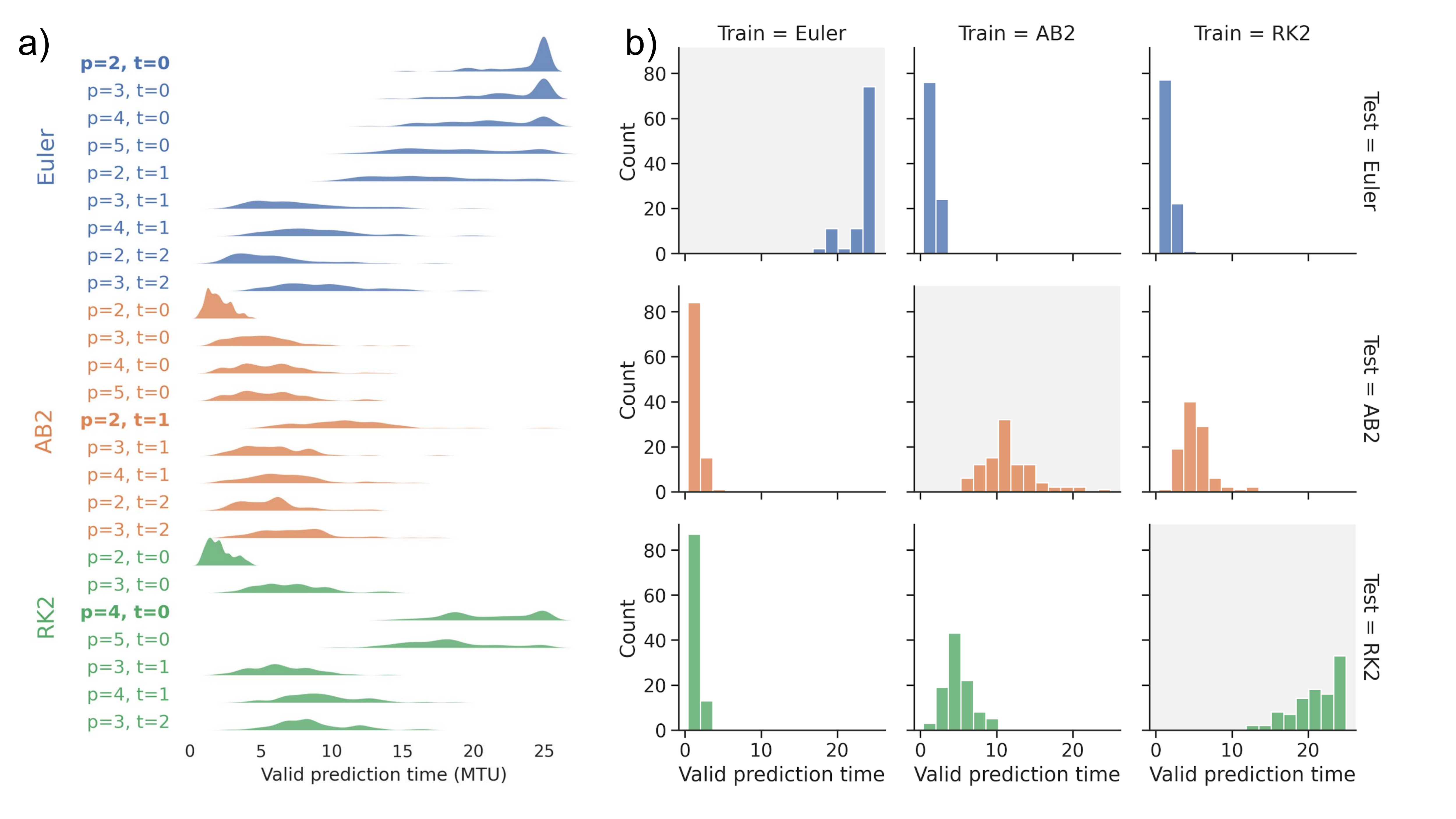}
    \caption{(a) Distribution of valid prediction time from the 100 test samples of L96-6D using NVAR-RC with different combinations of hyperparameters trained and tested against datasets generated by Euler (top), \nth{2} order Adam-Bashforth (AB2; middle), and \nth{2} order 2-stage Runge-Kutta (RK2; bottom) numerical integration schemes. The best performing combination of hyperparameters for each dataset is highlighted in bold. (b) Cross-validation of the best performing NVAR-RC (bolded selections in (a)) trained and tested using all combinations of the three different datasets. The histograms show the distribution of VPT for the 100 test samples. The NVAR-RC performs poorly when the numerical scheme used to generate the test data is inconsistent with the scheme used to generate the training data.}
    \label{fig:vpt_cross}
\end{figure}

Since the NVAR-RC has a deep connection to the underlying numerical integration scheme used for generating data, it is natural to examine how the best-performing if not optimal configurations of NVAR-RC changes with different numerical schemes and cross-validate an NVAR-RC optimized for one numerical scheme against data from another scheme. Figure \ref{fig:vpt_cross} (a) shows the distribution of the valid prediction time for L96-6D for different combinations of hyperparameters and numerical schemes. As shown earlier, the \nth{2} degree polynomial features with no time-lag performs the best for the Euler dataset. For \nth{2} order Adam-Bashforth (AB2) scheme, 1 time-lag is required for the best performance since the scheme itself uses one previous step for advancing. To best predict the \nth{2} order 2-stage Runge-Kutta (RK2) dataset, the 4th degree polynomial features are required, which is consistent to the corresponding time-stepping equations (not shown). It is worth noting that the best performance is achieved by providing the necessary features, and including more unnecessary features will allow for more error in the entries of bigger $\mathbf{W}_{out}$ that degrades the performance.

Figure \ref{fig:vpt_cross} (b) cross-validates the best performing NVAR-RC from (a) on predicting test data generated from different numerical schemes. It is clear that the valid prediction time is large (e.g., $>10 MTU$) when the training and testing numerical schemes are the same. In the case where the two are inconsistent, astonishingly poor performance appears. It has a general implication that data-driven methods could overfit the data in the sense that it learns the underlying numerical scheme and generalizes poorly to a dataset that uses other schemes.

\subsection{Sensitivity to training data length and noise}
\begin{figure}
    \centering
    \includegraphics[width=0.8\columnwidth]{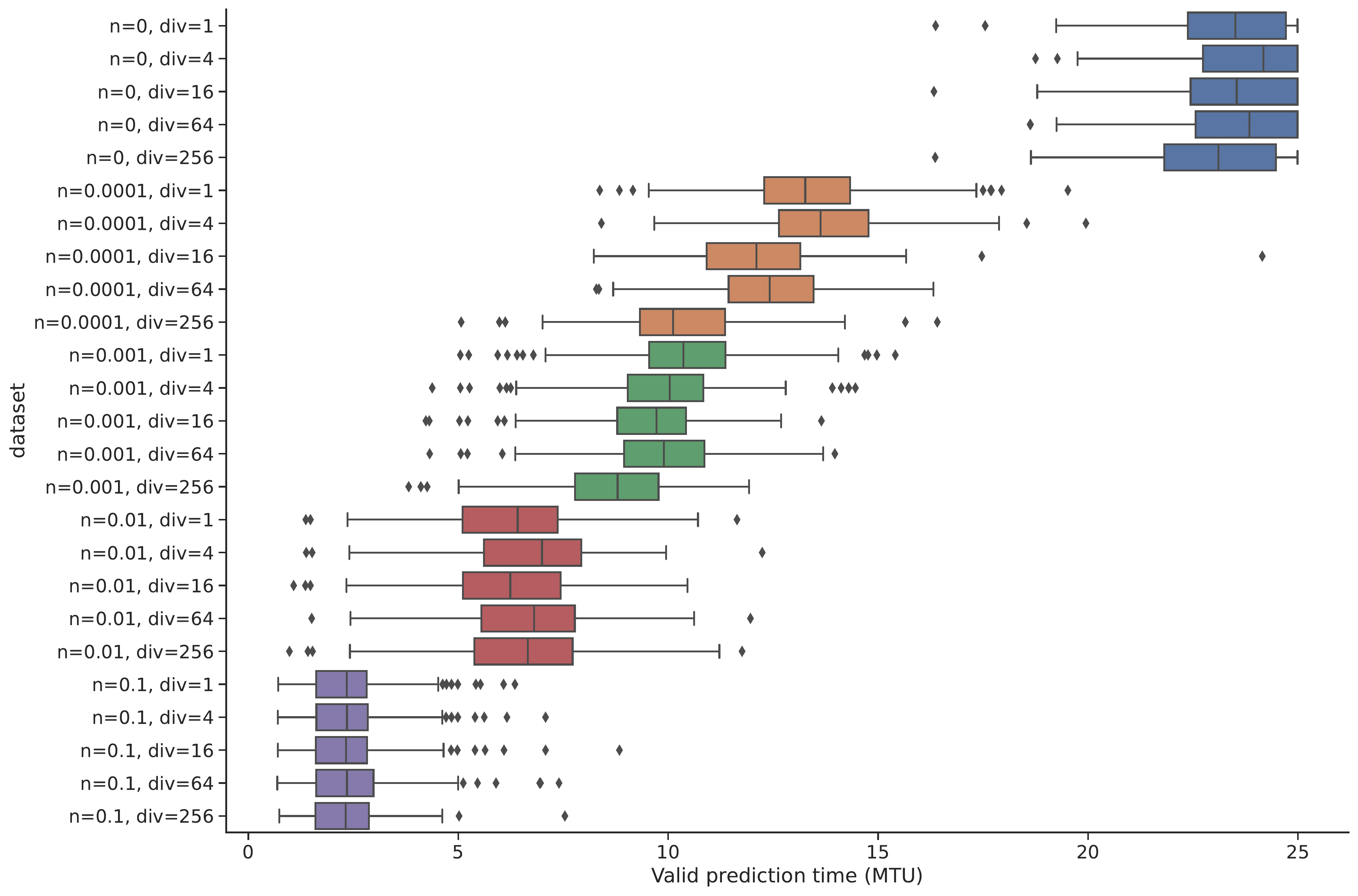}
    \caption{Comparisons of VPT of L63 with Euler scheme for different combinations of additive noise
magnitude $n$ and length of the training data. The training data length was reduced to the division $div$ of the total 4000 MTU (400,000 steps). The shortest training length tested was 15 MTU ($div=256$; 1,500 steps)}
    \label{fig:vpt_dlen_noise}
\end{figure}

Here we show the sensitivity of NVAR-RC performance for predicting L63 to the training data length and noise in Figure \ref{fig:vpt_dlen_noise}. From a total of 4000 MTU of training data, we gradually decrease the data length by division of 1, 4, 16, 64, and 256. For noise-free data, the training length does not seem to change the NVAR-RC performance, which is not surprising as the training process is merely fitting a small number of entries in $\mathbf{W}_{out}$ (e.g., 30 entries for L63 using the \nth{2} degree polynomial features with no time-lag) from a data length larger than at least 1500 time steps. As the additive noise level increases, the NVAR-RC VPT decreases from the almost perfect 23 MTU to around 5-15 MTU for smaller noises. For noise level of $0.1$, the VPT is only around 2 MTU. The results show that NVAR-RC reasonably degrades in the case of noisy data and is more or less insensitive to the training data length when it is significantly larger then the size of $\mathbf{W}_{out}$. 

\section{Importance of special functions: the Colpitts Oscillator example}
To demonstrate the importance of including special functions in addition to the polynomial features, we introduce the Colpitts Oscillator, an electronic oscillator circuit that exhibits chaotic behavior, in its dimensionless first-order differential form (see \citet{creveling_parameter_2008} and references therein):

\begin{equation}
\begin{aligned}
    \frac{dx_1}{dt} &= \alpha x_2 \\
    \frac{dx_2}{dt} &= -\gamma (x_1+x_3)-qx_2 \\
    \frac{dx_3}{dt} &= \eta(x_2+1-e^{-x_1}),
\end{aligned}
\end{equation}
where common choice of model parameters are $\alpha=5$, $\gamma=0.08$, $q=0.7$, and $\eta=6.3$. 

\begin{figure}
    \centering
    \includegraphics[width=\columnwidth]{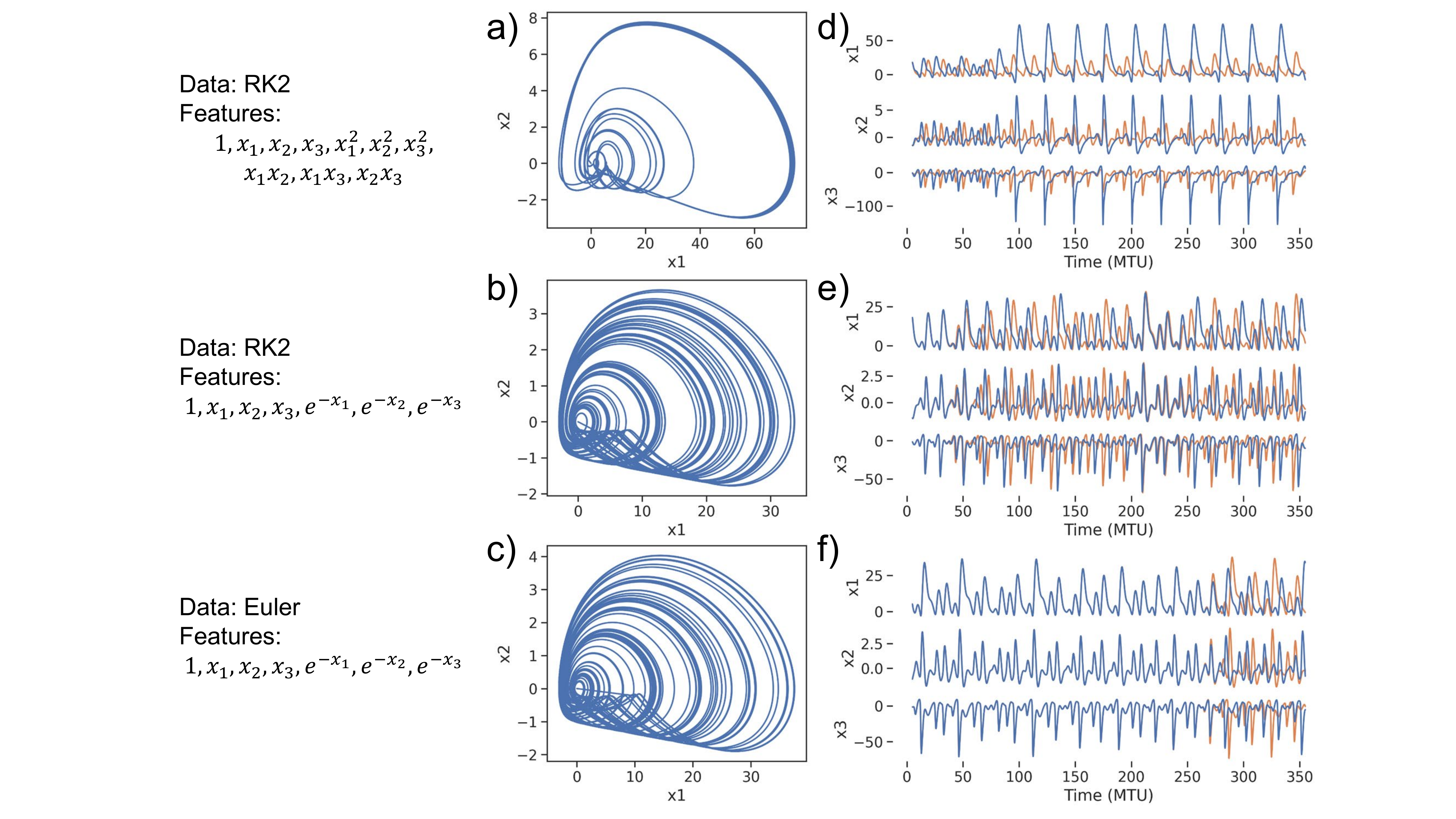}
    \caption{(a)-(c) Predicted attractors of the Colpitts Oscillator from NVAR-RCs trained from different datasets with different features in $\mathbf{R}$. Note that the preconditioning was necessary for including the \nth{2} degree polynomial terms and the exponential functions in the features. (d)-(f) True (orange) and predicted (blue) trajectories from the NVAR-RCs with different features and using datasets generated from different integration schemes. The training data length is 1000 MTU (100,000 steps; 72 Lyapunov time)}
    \label{fig:colpitts}
\end{figure}

Figure \ref{fig:colpitts} shows the performance of NVAR-RCs with different configurations on reconstruction of the attractor of Colpitts Oscillator and on the prediction in testing dataset. Using standard \nth{2} degree polynomial features, NVAR-RC trained from RK2 data yields an attractor that appears similar but with a larger scale. The prediction diverges from truth almost immediately. Intuitively, the poor performance is expected since there presents an exponential term on the right-hand side of the dynamical equation. At best, the NVAR-RC constructs a \nth{2} order approximation to the exponential term. The \nth{2} order approximation error builds up quickly over few iterations and lead to the poor performance. When replacing the \nth{2} degree polynomial features with the exponential of all variables, the resulting NVAR-RC gives attractor that appears indistinguishable from the truth to the human eye. The prediction tracks well with the true trajectory for about 50 MTU ($\sim$ 3.6 Lyapunov time). Switching to training and testing on the Euler dataset does not seem to change the attractor, but the valid prediction time is extended by a factor of 5 to beyond 250 MTU ($\sim$ 18 Lyapunov time). 

We like to note here that the preconditioning is especially important for successful computation of $\mathbf{W}_{out}$ since the variables oscillate with a magnitude of 20-50 and the exponential of the variables leads to severely ill-conditioned estimation of $\mathbf{W}_{out}$. Also, in this case, normalizing data changes the behavior of the exponential function as it is sensitive to the magnitude of the input and leads to a NVAR-RC with poor performance if converging at all. Another issue about generalizing the inclusion of special functions is the estimation of coefficients within the function, for example $c$ from $e^{cx}$. Here we avoid this problem by prescribing the $-1$ from $e^{-x}$ in Colpitts Oscillator equations. While it is possible to estimate the coefficients using gradient-descend methods, it is not clear how it can be done using the least-squares regression. 

Recall that \cite{gauthier_next_2021} demonstrated that NVAR-RC, without additional effort, learns well the double-scroll electronic circuit \citep{chang_stabilizing_1998} which also has a special function $sinh(x)$ in the governing equations. One major difference in this case is that the variables in the double-scroll system are bounded between -2 and 2 where $sinh(x)$ is in the linear regime and can be approximated by \nth{3} degree polynomial features used in their experiment.

\section{Empirical high-order time-stepping dynamical equations trained from skipped data}\label{SEC:skip}
In the previous section, we showed that NVAR-RC empirically discovers the dynamical equations in Euler time-stepping form without any prior knowledge about the chosen time integration scheme, when using a training dataset generated using the Euler numerical integration scheme. Motivated by the half-timestep approach commonly adopted in higher-order schemes, such as the Runge-Kutta methods, we explore whether NVAR-RC is able to construct an empirical high-order integration scheme in time-stepping form that predicts well the same data with a step size larger than multiple model integration time steps. In realistic applications, the temporal resolution of a stored output data that is available for training is usually larger than the timestep used by the numerical integration scheme (e.g., climate model simulations and reanalysis datasets). In addition, an important use case for ML emulators, including NVAR-RC, is to form a surrogate of the costly model in applications such as generating long climate simulations \citep[e.g.,][]{scher_toward_2018,scher_weather_2019} and estimating forecast uncertainty from a large ensemble of realizations \citep{penny_integrating_2021}. It would be useful for NVAR-RC to learn a high-order time-stepping representation with a step size larger than the model integration time step to reduce the run time. We note that it would be challenging for a human to design such a customized empirical high-order time-stepping scheme that matches the data frequency with a larger step size.

We examine this potential use case by training NVAR-RCs on the L63 dataset with skip sizes of 0, 1, 2, 4, and 8. The experimental setup is the same as before, except for the training data length. Here we consider the training process as a fitting problem of the $\mathbf{W}_{out}$ entries and try to compare fairly by making the training data length proportional to the number of features. We use 1 MTU of data for training each feature, for example, a 10-feature NVAR-RC would be trained with 10 MTU of data.

\begin{figure}
    \centering
    \includegraphics[width=\columnwidth]{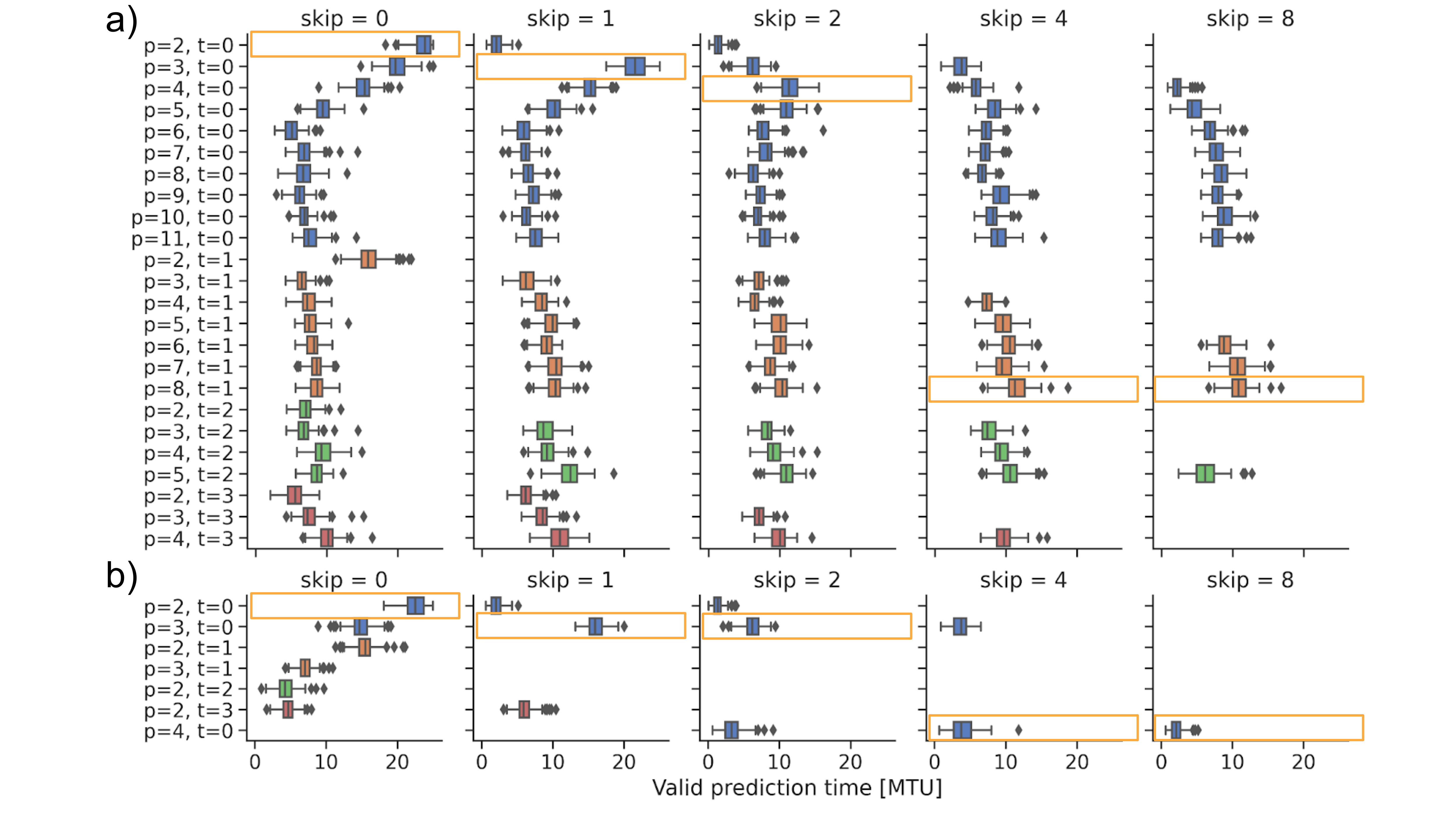}
    \caption{Comparisons of L63 VPT for different skip sizes using the best performing combinations of highest degree $p$ of polynomial features and the number of time-lags $t$ for training NVAR-RC (a) with preconditioning, and (b) without preconditioning. The optimized parameters for each skip size are highlighted with a yellow box. Entries for some combinations of $p$, $t$, and skip size are empty due to the failure of $\mathbf{W}_{out}$ computation. Note that good prediction (e.g., VPT>10 MTU) for larger skip size generally requires more features, but more features does not necessarily yield better performance.}
    \label{fig:vpt_skip}
\end{figure}

In figure \ref{fig:vpt_skip}, we show the results of a grid-search through different hyperparameter combinations of the highest degree $p$ of polynomial features and the number of time-lags $t$ for each skip size. The $p=2, t=0$ NVAR-RC shows the baseline for 0 skip size, which is the best performing NVAR-RC for the Euler dataset (\nth{2} degree polynomial features with 0 time-lag), predicting up to 24 MTU. It is intuitive that the VPT of the optimized NVAR-RC (highlighted with yellow boxes) decreases as the skip size increases, but in general the optimized VPT remains larger than 10 MTU for the preconditioned NVAR-RC. Without preconditioning, it is difficult to make use of the higher-degree polynomial features and the larger skip sizes experiment results in VPT less than 5 MTU. The optimized parameters show that the highest degree $p$ of polynomial features (and sometimes number of time-lags $t$) required for good performance also increases with the skip size, but further increasing $p$ and $t$ results in degradation. 

\section{Discussion}
We have identified the connection between the NVAR-RC $\mathbf{W}_{out}$ and the dynamical equations in time-stepping form, both of which are inevitably connected via the underlying numerical integration scheme. This has a broader implication that data-driven methods in general could overfit the training data in the sense that the resulting models could be specific to the numerical methods used to generate the data. When cross-validating the trained data-driven methods to make predictions with test data generated by different numerical schemes, the prediction skill degrades severely. This result gives a warning that one must be careful when developing data-driven models with simulated datasets (that are imprinted by the numerical scheme). Note that the issue is not specific to the NVAR-RC, but emerges as data-driven methods push the limit of VPT such that the underlying time integration scheme becomes important. However this property also presents an opportunity, in that the NVAR-RC empirically `learns' a numerical integration scheme that best fits the data taken from nature, which may be a challenge if designed by a human. 

It is worth-noting that the time-lag information plays two roles in NVAR-RC. The first role as noted widely in the literature is for generating an embedding of the attractor provided by Taken's embedding theorem \citep{takens_detecting_1981}. This role may be more important in the case of reconstructing a partially observed system where the information of the unmeasured states is contained in the time-lags. In addition, the time-lags also allows for NVAR-RC to learn multi-step integration schemes such as the \nth{2} order Adam-Bashforth scheme used in Figure \ref{fig:vpt_cross}. In this paper, the second role may be dominant as all the dynamical systems are fully observed. This is also indicated by the fact that the number of time-lags $t>0$ is only beneficial for the AB2 dataset and the skip-prediction task. 

NVAR-RC can be viewed as a mapping form of the SINDy algorithm, since both perform nonlinear vector autoregression (NVAR) from a functional library for dynamics identification. The only difference is that NVAR-RC (re)constructs the dynamical equations in time-stepping form while SINDy operates in differential form by approximating the derivatives from data. By drawing this connection, we have started to explore extending the NVAR-RC's functional library beyond polynomial features to include more special functions and transformations of features. One successful example was given by \cite{lin_fourier_2021}, where a traditional RC was trained in spectral space and successfully emulated a coupled atmosphere-ocean quasi-geostrophic model \citep{de_cruz_modular_2016}. As shown with the Colpitts Oscillator example, it can be beneficial to include all special functions in the target dynamical equations, which would require prior knowledge of the system. The special functions, if not provided, will be approximated empirically by the polynomial features in a style similar to Taylor expansion. This approximation will lead to a degraded surrogate representation of the system, but could still be useful in tasks like short-term prediction. Sometimes the approximation is sufficiently accurate that even the attractor can be reconstructed successfully as in the double-scroll example shown by \cite{gauthier_next_2021}. We note that a more general approach using a radial basis function (e.g., Gaussian) to account for nonlinearity beyond polynomial features is explored in \citet{clark_reduced_2021}. Because we expect the $\mathbf{W}_{out}$ matrix to be sparse by nature, for example due to the local influence of geophysical dynamics, the LASSO algorithm (as used for SINDy) could be used as a replacement for linear regression to identify the optimal $\mathbf{W}_{out}$.

For applications to high-dimensional spatially extended physical systems, a sparse $\mathbf{W}_{out}$ would be expected (given the correct features are included). In most classical dynamics (e.q., fluid dynamics), the contributing processes in short-term prediction are mostly local. It would be more accurate and efficient to apply localization to $\mathbf{W}_{out}$ (i.e. by explicitly setting the spatially distant interactions to zero, e.g. \citet{liu_model-free_2021}). In addition, a homogeneity assumption can be applied for systems with spatially symmetric processes. For example, the $\mathbf{W}_{out}$ of the L96-6D system is a concatenation of multiple diagonal matrices with shifts corresponding to the highly localized and symmetric diffusion and advection processes. As a result, the estimation of the $\mathbf{W}_{out}$ can be reduced to a single column vector (symmetry in space) with a smaller row size (locality in dynamics). The localized nature could also be viewed as analogous to the kernel of the convolutional neural network. \cite{bocquet_data_2019} provided similar discussion of the locality and homogeneity for inferring dynamics in differential form from data assimilation.

In this study, we examined to some extent how noise in the training data affects the NVAR-RC performance. The results show reasonable degradation in performance as noise level increases. A potential route for mitigating the negative impact of noisy data would be integrating NVAR-RC and data assimilation for simultaneously mitigating errors in the data while training the NVAR-RC model. Recently, we have shown that data assimilation can synchronize the standard RC using partial and noisy observations \citep{penny_integrating_2021}. A number of recent studies \citep[e.g.,][]{bocquet_data_2019,bocquet_bayesian_2020} have focused on joint estimation of state and surrogate model parameters from partial and noisy data for inferring the dynamical model in differential form. It may be worth exploring a similar approach using NVAR-RC for it represents not only the dynamical equations but the underlying time integration scheme.

Because the numerical scheme is also identified, the time-stepping equation inferred using NVAR-RC generally provides better prediction skill versus the identification of dynamics in differential form in the case when the identified surrogate model uses a different integration scheme than that which is used to generate the data. This apparent `benefit' in performance comes with a trade-off of reduced interpretability, in particular for high-order schemes. The right-hand side of the original dynamical equations in differential form becomes intractable in the time-stepping form when going from the \nth{1} order Euler scheme to higher-order schemes as more higher-degree polynomial features start to emerge. On the other hand, we showed that when training with skipped data (i.e., a coarser temporal resolution), the NVAR-RC was able to construct an empirical high-order time-stepping scheme with a step size several times larger than the numerical integration time step originally used to generate the data. It achieved this by using higher-degree polynomial features and time-lag information to result in quite astonishing prediction skill (VPT $>10$ MTU). There is a tradeoff between the number of iterations and the computational cost for a single iteration for this empirical skip-prediction. As opposed to some high-order multistage time integration schemes (e.g., Runge-Kutta methods) that have serial dependence on previous stages, such a skip-prediction could be desirable from the perspective of producing high performance surrogate models since the required computations are parallelizable matrix operations.

\section{Acknowledgements}
T.-C. Chen is supported by the NOAA Cooperative Agreement with CIRES, NA17OAR4320101. S.G. Penny and J.A. Platt acknowledge support from the Office of Naval Research (ONR) grants N00014-19-1-2522 and N00014-20-1-2580. S.G. Penny and T.A. Smith acknowledge support from NOAA grant NA20OAR4600277.

\section{Author contributions}
T.-C. Chen performed the simulations and conceptualized the connection between NVAR-RC and dynamical equations in time-stepping form. T.-C. Chen and S.G. Penny developed the concept of skip-prediction and drafted the manuscript. T.A. Smith and J.A. Platt helped interpret the results and improved the manuscript. Coauthors Chen, Penny, Smith, and Platt all contributed to the software used to produce the results.

\appendix

\FloatBarrier

\bibliographystyle{unsrtnat}
\bibliography{references.bib}  






\end{document}